\documentclass[11pt]{article}
\usepackage{acl2014}
\usepackage{times}
\usepackage{url}
\usepackage{latexsym}
\usepackage{xspace}
\usepackage{booktabs}
\usepackage{color}
\usepackage{graphicx}
\usepackage{tabulary}
\usepackage{enumitem}
\usepackage{amsfonts}

  {\begin{itemize}[topsep=0pt, partopsep=0pt] %
    \setlength{\itemsep}{0pt}%
    \setlength{\parskip}{0pt}%
    }%
  {\end{itemize}}
  
  \usepackage{color}
  {\begin{enumerate}≈ \footnotesize%
    \setlength{\itemsep}{0pt}%
    \setlength{\parskip}{0pt}%
    }%
  {\end{enumerate}}

\usepackage{microtype}
\usepackage{multirow}
\usepackage{verbatim}
\usepackage{amsmath,amsthm,amssymb}
\usepackage{array}
\usepackage[scaled=0.86]{helvet}
\usepackage{ifthen}
\usepackage{courier}
\usepackage[linesnumbered,vlined,ruled]{algorithm2e}

\newcommand{\captionfonts}{\small}
\makeatletter  
\long\def\@makecaption#1#2{%
  \vskip\abovecaptionskip
  \sbox\@tempboxa{{\captionfonts #1: #2}}%
  \ifdim \wd\@tempboxa >\hsize
    {\captionfonts #1: #2\par}
  \else
    \hbox to\hsize{\hfil\box\@tempboxa\hfil}%
  \fi
  \vskip\belowcaptionskip}
\makeatother   

\belowdisplayskip \abovedisplayskip

\DeclareMathOperator*{\argmax}{arg\,max}

\newcommand{\eos}{{\it EOS}\xspace}
\newcommand{\sts}{{{\textsc{Seq2Seq}}}\xspace}

\title{Modeling Mutual Information and Promoting Diversity in Neural Machine Translation}
\title{Modeling Mutual Information and Promoting Diversity in Improving Neural Machine Translation
with a Mutual Information Objective and A Diversity-Promoting Decoder}
\title{Modeling Mutual Information in Neural Machine Translation}
\title{Mutual Information and Diverse Decoding Improve Neural Machine Translation}

\author{Jiwei Li and Dan Jurafsky \\
Computer Science Department \\
{Stanford University, Stanford, CA, 94305, USA} \\
{\tt jiweil,jurafsky@stanford.edu} \\
}
\date{}

\begin{document}
\maketitle

\begin{abstract}
Sequence-to-sequence neural translation models
learn semantic and syntactic relations between sentence
pairs by optimizing the likelihood of the target given the source, i.e., $p(y|x)$,
an objective that ignores other potentially useful sources of information.
We introduce an alternative objective function for neural MT that maximizes the mutual information 
between the source and target sentences,
modeling the bi-directional dependency of sources and targets.
We implement the model with a simple re-ranking method,
and also introduce a decoding algorithm that increases diversity in the N-best
list produced by the first pass.
Applied to the WMT German/English and French/English tasks,
the proposed models offers a consistent performance boost
on both standard LSTM and attention-based neural MT architectures.
\end{abstract}
\section{Introduction}
Sequence-to-sequence models for machine translation (\sts) \cite{sutskever2014sequence,bahdanau2014neural,cho2014learning,kalchbrenner2013recurrent,sennrich2015improving,sennrich2015neural,gulcehre2015using} are of growing interest  for their 
capacity to learn semantic and
syntactic relations between sequence pairs, capturing contextual
dependencies in a more continuous way than phrase-based SMT approaches. 
\sts models require minimal domain knowledge, can be trained
end-to-end, have a much smaller memory footprint than the
large phrase tables needed for phrase-based SMT,
and achieve  state-of-the-art performance in large-scale tasks like English to French \cite{luong2015addressing} and English to German  \cite{luong2015effective,jean2014using} translation.

\sts models are implemented as an encoder-decoder network, in which
a source sequence input $x$ is mapped (encoded) to a continuous  vector representation
from which  a target output $y$ will be generated (decoded).
The framework 
is optimized through maximizing the log-likelihood of observing the paired output  $y$ given $x$:
\begin{equation}
\text{Loss}= -\log p(y|x)
\label{eqseq2seq}
\end{equation}
While standard \sts models thus capture the unidirectional dependency from source to target, i.e., $p(y|x)$,
they ignore $p(x|y)$, the dependency from the target to the source,
which has long been an important feature in phrase-based translation 
\cite{OchNey02,shen2010string}.
Phrase based systems that combine $p(x|y)$, $p(y|x)$ and other features
like sentence length yield significant performance boost. 


%

We propose to incorporate this bi-directional dependency and model the maximum mutual information (MMI) between source and target
into  \sts models.
%
As \newcite{li2015diversity} recently showed in the context of
conversational response generation, 
the MMI based objective function 
 is equivalent
to linearly combining $p(x|y)$ and $p(y|x)$.
With a tuning weight $\lambda$, 
such a loss function can be written as :
\begin{equation}
\begin{aligned}
\hat{y}&=\argmax_{y} \log\frac{p(x,y)}{p(x)p(y)^{\lambda}} \\
&= \argmax_{y} (1-\lambda) \log p(y|x)+\lambda \log p(x|y) 
\label{eqseq2seq}
\end{aligned}
\end{equation}
But as also discussed in \newcite{li2015diversity}, direct decoding from \eqref{eqseq2seq} 
is infeasible because computing $p(x|y)$ cannot be done until the 
target has been computed\footnote{
As demonstrated in \cite{li2015diversity}
\begin{equation}
\log\frac{p(x,y)}{p(x)p(y)^\lambda}=\log p(y|x)-\lambda \log p(y)
\end{equation}
Equ. \ref{eqseq2seq} can be immediately achieved by applying bayesian rules 
$$\log p(y)=\log p(y|x)+\log p(x)-\log p(x|y)$$
}. 

To avoid this enormous search space, we propose
to use a reranking approach to approximate
the mutual information between source and target
in neural machine translation models.
We separately trained two \sts models, one for $p(y|x)$ and one for $p(x|y)$.
The $p(y|x)$ model is used to
generate N-best lists from the source sentence $x$.  The lists are followed by a reranking process 
using the second term of the
objective function, $p(x|y)$. 

Because reranking approaches are dependent on having a diverse
N-best list to rerank, we also propose a diversity-promoting decoding model
tailored to neural MT systems. 
We tested the  mutual information objective function and the diversity-promoting
decoding model on English$\rightarrow$French, English$\rightarrow$German
and German$\rightarrow$English  
 translation tasks,
using both standard LSTM settings and the more advanced attention-model based settings
that have recently shown to result in higher performance.
 
The next section presents related work, followed by a background section 3 introducing 
LSTM/Attention machine translation models. Our proposed model will be described in detail in Sections 4,
with datasets and experimental results in Section 6 followed by conclusions.

\section{Related Work}

This paper draws on three prior lines of research: \sts models, modeling mutual information,
and promoting translation diversity.

\paragraph{\sts Models} 
\sts models map source sequences to vector space representations, 
from which a target sequence is then generated. They yield good
performance in a variety of NLP generation tasks including 
conversational response generation
\cite{vinyals2015neural,serban2015building,li2015diversity}, and
parsing \cite{vinyals2014grammar,luong2015multi}.

A neural machine translation system 
uses distributed representations to
model the conditional probability of targets given sources,
using two components, an encoder and a decoder. 
Kalchbrenner and Blunsom \shortcite{kalchbrenner2013recurrent} used an encoding model akin to convolutional networks for encoding
and standard hidden unit recurrent nets for decoding.
Similar convolutional networks are used in \cite{meng2015encoding} for encoding. 
\newcite{sutskever2014sequence,luong2015effective} employed a stacking LSTM model for both encoding and decoding. 
\newcite{bahdanau2014neural}, \newcite{jean2014using} adopted bi-directional recurrent nets for the encoder. 

\paragraph{Maximum Mutual Information}
Maximum Mutual Information (MMI) was introduced in speech recognition \cite{bahl1986maximum}
as a way of measuring
the mutual dependence between inputs (acoustic feature vectors) and outputs (words)
and improving discriminative training \cite{WoodlandPovey02}.
\newcite{li2015diversity} show that MMI could solve
an important problem in \sts conversational response generation.
Prior \sts models tended to generate highly generic, dull responses (e.g., {\em I don't know}) regardless of the inputs
\cite{sordoni2015neural,vinyals2015neural,serban2015survey}.  Li
et al. \shortcite{li2015diversity} show that modeling
the mutual dependency between messages and response promotes the
diversity of response outputs.

Our goal, distinct from these previous uses of MMI,
is to see whether the mutual information objective
improves translation by bidirectionally modeling source-target dependencies.
In that sense, our work is designed to incorporate into \sts models
features that have proved useful in phrase-based MT,
like the reverse translation probability or sentence length
\cite{OchNey02,shen2010string,devlin2014fast}.

\paragraph{Generating Diverse Translations}

Various algorithms have been proposed for generated diverse translations in phrase-based MT,
including compact representations like lattices and hypergraphs
\cite{macherey2008lattice,tromble2008lattice,kumar2004minimum},
``traits'' like translation length 
\cite{devlin2012trait}, bagging/boosting
\cite{xiao2013bagging}, or multiple systems \cite{cer2013positive}.
\newcite{gimpel2013systematic,batra2012diverse}, 
produce diverse N-best lists by adding a dissimilarity function
based on N-gram overlaps, distancing the current translation from already-generated ones
by choosing translations that have higher scores but distinct from previous ones.
While we draw on these intuitions, these
existing diversity promoting algorithms are tailored to phrase-based translation frameworks and not easily transplanted to neural MT decoding which requires batched computation.  
\section{Background: Neural Machine Translation} 
Neural machine translation models map source $x=\{x_1,x_2,...x_{N_x}\}$ to a continuous  vector representation, from which 
target output $y=\{y_1,y_2,...,y_{N_y}\}$ is to be generated. 

\subsection{LSTM Models}
 A long-short term memory model \cite{hochreiter1997long}
 associates each time step with an input gate, a memory gate and an output gate, 
denoted respectively as $i_t$, $f_t$ and $o_t$.
Let $e_{t}$ denote the vector for the current word $w_t$, $h_t$ the vector computed by the LSTM model at time $t$ by combining $e_t$ and $h_{t-1}$.,
$c_t$ the cell state vector at time $t$,
and $\sigma$ the sigmoid function. The vector representation $h_t$ for each time step $t$ is given by:
\begin{eqnarray}
i_t=\sigma (W_i\cdot [h_{t-1},e_{t}])\\
f_t=\sigma (W_f\cdot [h_{t-1},e_{t}])\\
o_t=\sigma (W_o\cdot [h_{t-1},e_{t}])\\
l_t=\text{tanh}(W_l\cdot [h_{t-1},e_{t}])\\
c_t=f_t\cdot c_{t-1}+i_t\cdot l_t\\
h_{t}^s=o_t\cdot \text{tanh}(c_t)
\end{eqnarray}
where $W_i$, $W_f$, $W_o$, $W_l \in \mathbb{R}^{K\times 2K}$.
The LSTM defines a distribution over outputs $y$ and sequentially predicts tokens using a softmax function:
\begin{equation*}
\begin{aligned}
p(y|x)=\prod_{t=1}^{n_T}\frac{\exp(f(h_{t-1},e_{y_t}))}{\sum_{w'}\exp(f(h_{t-1},e_{w'}))}
\end{aligned}
\label{equ-lstm}
\end{equation*}
where $f(h_{t-1}, e_{y_t})$ denotes the activation function between $h_{t-1}$ and $e_{w_t}$, where $h_{t-1}$ is the representation output from the LSTM at time \mbox{$t-1$}. 
Each sentence concludes with a special end-of-sentence symbol \eos. 
Commonly, the input and output each use different LSTMs with separate sets of compositional parameters to capture different compositional patterns. 
During decoding, the algorithm terminates when an \eos token is predicted.
\subsection{Attention Models}
Attention models adopt a look-back strategy that
links the current decoding stage with input time steps
to represent which portions of the
input are most responsible for the current decoding state \cite{xu2015show,luong2015addressing,bahdanau2014neural}. 

Let $H=\{\hat{h}_1,\hat{h}_2,...,\hat{h}_{N_x} \}$ be the collection of hidden vectors outputted from LSTMs during encoding.  
Each element in $H$ contains information
about the input sequences, focusing
on the parts surrounding each specific token. 
Let  $h_{t-1}$ be the LSTM outputs for decoding at time $t-1$. 
Attention models link the current-step decoding information, i.e., $h_{t}$ with each of the 
representations at decoding step $\hat{h}_{t'}$ using a weight variable $a_t$.
$a_t$ can be constructed from different scoring functions such as
the {\it  dot product} between the two vectors, i.e., $h_{t-1}^T\cdot \hat{h}_t$, a {\it general} model akin to tensor operation i.e., $h_{t-1}^T\cdot W\cdot \hat{h}_t$, and the {\it concatenation} model by concatenating the two vectors i.e., $U^T\cdot$\text{tanh}$(W\cdot [h_{t-1}, \hat{h}_t]$). 
The behavior of different attention scoring functions have been extensively studied in \newcite{luong2015effective}.
For all experiments in this paper, we adopt the {\it general} strategy where the relevance score between 
the current step of the decoding representation and the encoding representation is given by:
\begin{equation}
\begin{aligned}
&v_{t'}=h_{t-1}^T\cdot W\cdot \hat{h}_t\\
&a_i=\frac{\exp(v_{t^*})}{\sum_{t*}\exp(v_{t^*})}\\
\end{aligned}
\end{equation}
The attention vector  is created by averaging weights over all input time-steps: 
\begin{equation}
m_t=\sum_{t'\in[1,N_S]}a_i \hat{h}_{t'}
\end{equation}
Attention models predict subsequent tokens based on the combination of the last step outputted LSTM vectors $h_{t-1}$ and attention vectors $m_t$:
\begin{equation}
\begin{aligned}
&\vec{h}_{t-1}=\text{tanh}(W_c\cdot [h_{t-1}, m_t])\\
&p(y_t|y_<,x)=\text{softmax}(W_s\cdot\vec{h}_{t-1})
\end{aligned}
\end{equation}
where $W_c \in \mathbb{R}^{K\times 2K}$, $W_s\in \mathbb{R}^{V\times K}$ with V denoting vocabulary size. 
\newcite{luong2015effective} reported a significant performance boost by integrating $\vec{h}_{t-1}$ into the next step LSTM hidden state computation (referred to as the {\it input-feeding} model), making LSTM compositions in decoding as follows:
\begin{equation}
\begin{aligned}
&i_t=\sigma (W_i\cdot [h_{t-1},e_{t},\vec{h}_{t-1}])\\
&f_t=\sigma (W_f\cdot [h_{t-1},e_{t},\vec{h}_{t-1}])\\
&o_t=\sigma (W_o\cdot [h_{t-1},e_{t},\vec{h}_{t-1}])\\
&l_t=\text{tanh}(W_l\cdot [h_{t-1},e_{t},\vec{h}_{t-1}])\\
\end{aligned}
\end{equation}
where $W_i$, $W_f$, $W_o$, $W_l \in \mathbb{R}^{K\times 3K}$.
For the attention models implemented in this work, we adopt the {\it input-feeding} strategy.

 \subsection{Unknown Word Replacements}
One of the major issues in neural MT models is the computational complexity of the softmax function for target word prediction, which requires summing over all tokens in the vocabulary. 
Neural models tend to keep a shortlist of 50,00-80,000 most frequent words and use an unknown ({\tt UNK}) token to represent all infrequent tokens, which significantly impairs BLEU scores. Recent work has proposed to deal with this issue:
\cite{luong2015addressing} adopt a post-processing strategy based on aligner from IBM models, while
\cite{jean2014using} approximates softmax functions by selecting a small subset of target vocabulary. 

In this paper, we use a strategy similar to that of \newcite{jean2014using}, thus avoiding the reliance 
on external IBM model word aligner. 
From the attention models, we obtain word alignments from the training dataset, 
from which a bilingual dictionary is extracted. 
At test time, we first generate target sequences. Once a translation is generated, we link the generated {\tt UNK} tokens back to positions in the source inputs, and replace each {\tt UNK} token with the  translation word of 
its correspondent source token using the pre-constructed dictionary. 

As the unknown word replacement mechanism relies on automatic word alignment extraction which is not explicitly modeled in vanilla \sts models, it can not be immediately applied to vanilla \sts models.  However, since unknown word replacement can  be viewed as a post-processing technique, we can apply a pre-trained attention-model to any given translation. For \sts models, we first generate translations and replace {\tt UNK} tokens within the translations using the pre-trained attention models to post-process the translations. 

\section{Mutual Information via Reranking}
\begin{figure*}
\includegraphics[width=2in]{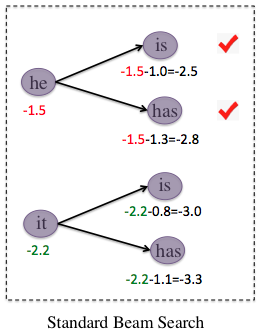}
\includegraphics[width=2.6in]{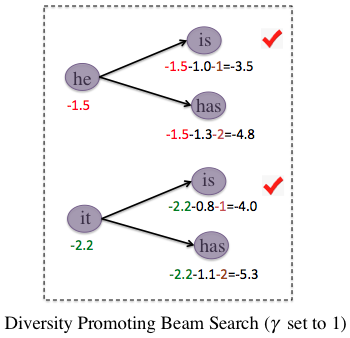}
\centering
\caption{Illustration of Standard Beam Search and proposed diversity promoting Beam Search. }
\end{figure*}
As discussed in \newcite{li2015diversity},
direct decoding from \eqref{eqseq2seq} is infeasible since  the second part, $p(x|y)$,
requires completely generating the target before it can be computed. 
We therefore use an approximation approach:
\begin{enumerate}
\item Train $p(y|x)$ and $p(x|y)$ separately using vanilla \sts models or Attention models.
\item Generate N-best lists from $p(y|x)$.
\item Rerank the N-best list by linearly adding $p(x|y)$. 
\end{enumerate}


\subsection{Standard Beam Search for N-best lists} 

N-best lists  are generated using a beam search decoder with beam size set to $K=200$ from $p(y|x)$ models.
As illustrated in Figure 1, at time step $t-1$ in decoding, we keep record of $N$ hypotheses based on score $S(Y_{t-1}|x)=\log p(y_1,y_2,...,y_{t-1}|x)$. As we move on to time step $t$, 
we expand each of the K hypotheses
(denoted as $Y_{t-1}^k=\{y_1^k,y_2^k,...,y_{t-1}^k\}$, $k\in [1,K]$), 
 by selecting top $K$  of the translations, denoted as  $y_t^{k,k'}$, $k'\in [1,K]$,  
  leading to the construction of $K\times K$ new hypotheses:
  $$[Y_{t-1}^k, y_t^{k,k'}], k\in [1,K], k'\in [1,K]$$
The score for each of the $K\times K$ hypotheses is computed as follows:
\begin{equation}
S(Y_{t-1}^k,y_t^{k,k'}|x)=S(Y_{t-1}^k|x)+\log p(y_t^{k,k'}|x,Y_{t-1}^k)
\end{equation}
In a standard beam search model, the top $K$ 
hypotheses are selected
(from the $K\times K$  hypotheses computed in the last step) based on 
the score $S(Y_{t-1}^k,y_t^{k,k'}|x)$. The remaining hypotheses are ignored as we proceed to the next time step. 

We set the minimum length and  maximum length to 0.75 and 1.5 times the length of sources. 
Beam size N is set to 200.
To be specific, at each time step of decoding, we are presented with $K\times K$ word candidates. 
We first add all hypotheses with an \eos token being generated at current time step to the N-best list. 
Next we preserve the top K unfinished hypotheses and move to next time step. 
We therefore maintain batch size of 200 constant when some hypotheses  are completed and taken down by adding in more unfinished 
hypotheses. This will lead the size of final N-best list for each input much larger than the beam size\footnote{For example, for the development set of the English-German WMT14 task, each input has an average of 2,500 candidates in the N-best list. }.

\subsection{Generating a Diverse N-best List}
Unfortunately, the N-best lists outputted from standard beam search are a poor surrogate 
for the entire search space \cite{finkel2006solving,huang2008forest}. 
The beam search algorithm can only keep a small proportion of candidates in the search space and
most of the generated translations in N-best list are similar, differing  only by punctuation
or minor morphological variations, with most of the words overlapping. 
Because this lack of diversity in the N-best list will significantly decrease the impact of 
our reranking process,  it is important to find a way to  generate a more diverse N-best list.

We propose  to change
the way $S(Y_{t-1}^k,y_t^{k,k'}|x)$ is computed in an attempt to promote diversity,
as shown in Figure 1. 
For each of the  hypotheses $Y_{t-1}^k$ ($he$ and $it$),  
we generate the top $K$ translations, 
 $y_t^{k,k'}$, $k'\in [1,K]$ as in the standard beam search model.
Next we  rank the $K$ translated tokens generated from the same parental hypothesis
based on $p(y_t^{k,k'}|x,Y_{t-1}^k)$ 
in descending order: {\it he is} ranks the first among {\it he is} and {\it he has}, and {\it he has} ranks second; 
similarly for {\it it is} and {\it it has}.
 
Next we rewrite the score for $[Y_{t-1}^k, y_t^{k,k'}]$ by adding an additional part $\gamma k'$, where $k'$ denotes the ranking of the current hypothesis among its siblings, 
which is first for {\it he is} and {\it it is}, second for {\it he has} and {\it it has}. 
\begin{equation}
\hat{S}(Y_{t-1}^k,y_t^{k,k'}|x)=S(Y_{t-1}^k,y_t^{k,k'}|x)-\gamma k'
\end{equation}
The top $K$ hypothesis are selected based on $\hat{S}(Y_{t-1}^k,y_t^{k,k'}|x)$ as we move on to the next time step. 
By adding the additional term $\gamma k'$, 
the model punishes bottom ranked hypotheses among siblings (hypotheses descended from the same parent).
When we compare newly generated hypotheses descended from different ancestors, the model gives more credit to  top hypotheses from each of different ancestors. 
For instance, even though the original score for {\it it is} is lower than {\it he has},
the model favors the former as the latter is more severely punished by the intra-sibling ranking part $\gamma k'$. 
The model thus generally favors choosing hypotheses from diverse parents, leading to a more diverse N-best list. 

The proposed model is straightforwardly implemented with minor adjustment to the standard beam search 
model\footnote{Decoding for neural based MT model using large batch-size can be expensive resulted from softmax word prediction function. The proposed model supports batched decoding using GPU, significantly speed up decoding process than other diversity fostering models tailored to phrase based MT systems. }. 

We employ the diversity evaluation metrics in \cite{li2015diversity}
to evaluate the degree of diversity of the N-best lists:
calculating the average number of distinct unigrams {\it  distinct-1} and bigrams {\it  distinct-2} in the N-best list given each source sentence, 
scaled by the total number of tokens. 
By employing the diversity promoting model with $\gamma$ tuned from the development set based on BLEU score,  the value of {\it  distinct-1}  increases from $0.54\%$ to 
$0.95\%$, and {\it distinct-2} increases from $1.55\%$ to $2.84\%$ for English-German translation. Similar phenomenon are observed from English-French translation tasks and details are omitted for brevity. 

\subsection{Reranking}
The generated N-best list is then reranked  by linearly combining $\log p(y|x)$ with $\log p(x|y)$. 
The score of the source given each generated translation can be immediately computed from the previously trained $p(x|y)$. 

Other than $\log p(y|x)$, we also consider $\log p(y)$, which denotes the average language model probability trained from monolingual data. It is worth nothing that integrating 
$\log p(y|x)$ and $\log p(y)$ into reranking is not a new one and has long been employed by in noisy channel models in standard MT. 
In neural MT literature, recent progress has demonstrated the effectiveness of modeling reranking with language model  \cite{gulcehre2015using}.

We also consider an additional term that takes into account the length of targets 
(denotes as $L_T$) in decoding.
We thus linearly combine the three parts, making the final ranking score for a given target candidate $y$ as follows:
 \begin{equation}
 \begin{aligned}
& Score(y)=\log p(y|x)+\lambda \log p(x|y)\\
&~~~~~~~~~~~~~~~~~~~+\gamma\log p(y)+ \eta L_T
 \end{aligned}
 \end{equation}
 We optimize $\eta$, $\lambda$ and $\gamma$ using MERT \cite{och2003minimum}
 BLEU score \cite{papineni2002bleu} on the development set. 
 
\section{Experiments} 
Our models are trained on the WMT'14 training dataset containing 4.5 million pairs for English-German 
and German-English translation, and 12 million pairs for English-French translation.
For English-German translation, we limit our vocabularies to
the top 50K most frequent words for both languages. For English-French translation, we keep 
the top 200K most frequent words for the source language and 80K for the target language. 
Words that are not in the vocabulary list are noted as the universal unknown token. 

For the English-German and English-German translation, we use newstest2013 (3000 sentence pairs) as the development set and translation performances are reported in BLEU  \cite{papineni2002bleu} on newstest2014 (2737) sentences. 
For English-French translation, we concatenate news-test-2012 and news-test-2013 
to make a development set (6,003 pairs in total) and 
evaluate the models on news-test-2014 with 3,003 pairs\footnote{As in \cite{luong2015effective}. All texts are tokenized with tokenizer.perl and
BLEU scores are computed with multi-bleu.perl}.
\subsection{Training Details for $p(x|y)$ and $p(y|x)$}
We  trained neural models on Standard \sts Models and  Attention Models.
We trained $p(y|x)$ following the standard training protocols described in \cite{sutskever2014sequence}.
$p(x|y)$ is trained identically but with sources and targets swapped. 

We adopt a deep structure with four LSTM
layers for encoding and four LSTM layers for decoding,
each of which consists of a different set of parameters. 
We followed the detailed protocols from \newcite{luong2015effective}:
each LSTM layer consists of 1,000 hidden
neurons, and the dimensionality of word embeddings
is set to 1,000. Other training details include: 
LSTM parameters and word embeddings are
initialized from a uniform distribution between [-0.1,0.1]; 
For English-German translation, we run 12 epochs in total. 
After 8 epochs, we start halving the learning rate after each epoch;
for English-French translation, the total number of epochs is set to 8,
and we start halving the learning rate after 5 iterations. 
Batch size is set to 128; gradient clipping is adopted by scaling gradients
when the norm exceeded a threshold of 5. Inputs are reversed. 

Our implementation on a single GPU\footnote{Tesla K40m, 1 Kepler GK110B, 2880 Cuda cores.}  processes 
approximately 800-1200 tokens per second. 
Training for the English-German dataset (4.5 million pairs) takes roughly 12-15 days.
For the French-English dataset, comprised of 12 million pairs, training takes roughly 4-6 weeks. 

\begin{table*}[!ht]
\centering
\begin{tabular}{lll}\hline
Model&Features&BLEU scores \\ \hline
Standard&p(y$|$x)& 13.2 \\
Standard&p(y$|$x)+Length&13.6 (+0.4)\\
Standard&p(y$|$x)+p(x$|$y)+Length & 15.0 (+1.4)\\
Standard&p(y$|$x)+p(x$|$y)+p(y)+Length& 15.4 (+0.4)\\
Standard&p(y$|$x)+p(x$|$y)+p(y)+Length+Diver decoding& 15.8 (+0.4)\\\hline
& & +2.6 in total \\\hline\hline
Standard+{\it UnkRep}&p(y$|$x)&14.7 \\
Standard+{\it UnkRep}&p(y$|$x)+Length&15.2 (+0.7) \\
Standard+{\it UnkRep}&p(y$|$x)+p(x$|$y)+Length&16.3 (+1.1) \\
Standard+{\it UnkRep}&p(y$|$x)+p(x$|$y)+p(y)+Length&16.7 (+0.4) \\
Standard+{\it UnkRep}&p(y$|$x)+p(x$|$y)+p(y)+Length+Diver decoding&17.3 (+0.3) \\\hline\hline
& & +2.6 in total \\\hline\hline
Attention+{\it UnkRep}&p(y$|$x)& 20.5 \\
Attention+{\it UnkRep}&p(y$|$x)+Length&20.9 (+0.4)\\
Attention+{\it UnkRep}&p(y$|$x)+p(x$|$y)+Length & 21.8 (+0.9)\\
Attention+{\it UnkRep}&p(y$|$x)+p(x$|$y)+p(y)+Length& 22.1 (+0.3)\\
Attention+{\it UnkRep}&p(y$|$x)+p(x$|$y)+p(y)+Length+Diver decoding& 22.6 (+0.3)\\\hline
& & +2.1 in total \\\hline\hline
\multicolumn{2}{c}{Jean et al., 2015 (without {\it ensemble})}&19.4\\
\multicolumn{2}{c}{Jean et al., 2015 (with {\it ensemble})}&21.6\\\hline
\multicolumn{2}{c}{\newcite{luong2015effective}  (with {\it UnkRep}, without {\it ensemble})}&20.9\\
\multicolumn{2}{c}{\newcite{luong2015effective}  (with {\it UnkRep}, with {\it ensemble})}&23.0\\\hline\hline
 \end{tabular}
\caption{BLEU scores from different models for on WMT14 English-German results. 
{\it UnkRep} denotes applying unknown word replacement strategy. 
{\it diversity} indicates diversity-promoting model for decoding being adopted. 
Baselines performances are reprinted from Jean {\em et al}. (2014),  Luong {\em et al}. 2015a.}
\label{tab1}
\end{table*}

\begin{table*}[!ht]
\centering
\begin{tabular}{lll}\hline
Model&Features&BLEU scores \\ \hline
Standard&p(y$|$x)& 29.0 \\
Standard&p(y$|$x)+Length&29.7 (+0.7)\\
Standard&p(y$|$x)+p(x$|$y)+Length & 31.2 (+1.5)\\
Standard&p(y$|$x)+p(x$|$y)+p(y)+Length& 31.7 (+0.5)\\
Standard&p(y$|$x)+p(x$|$y)+p(y)+Length+Diver decoding& 32.2 (+0.5)\\\hline
& & +3.2 in total \\\hline\hline
Standard+{\it UnkRep}&p(y$|$x)&31.0 \\
Standard+{\it UnkRep}&p(y$|$x)+Length&31.5 (+0.5) \\
Standard+{\it UnkRep}&p(y$|$x)+p(x$|$y)+Length&32.9 (+1.4) \\
Standard+{\it UnkRep}&p(y$|$x)+p(x$|$y)+p(y)+Length&33.3 (+0.4) \\
Standard+{\it UnkRep}&p(y$|$x)+p(x$|$y)+p(y)+Length+Diver decoding&33.6 (+0.3) \\\hline
& & +2.6 in total \\\hline\hline
Attention+{\it UnkRep}&p(y$|$x)& 33.4 \\
Attention+{\it UnkRep}&p(y$|$x)+Length&34.3 (+0.9)\\
Attention+{\it UnkRep}&p(y$|$x)+p(x$|$y)+Length & 35.2 (+0.9)\\
Attention+{\it UnkRep}&p(y$|$x)+p(x$|$y)+p(y)+Length& 35.7 (+0.5)\\
Attention+{\it UnkRep}&p(y$|$x)+p(x$|$y)+p(y)+Length+Diver decoding& 36.3 (+0.4)\\\hline
& & +2.7 in total \\\hline\hline
\multicolumn{2}{c}{LSTM (Google) (without ensemble))}&30.6\\
\multicolumn{2}{c}{LSTM (Google) (with ensemble)}&33.0\\\hline
\multicolumn{2}{c}{\newcite{luong2015addressing}, {\it UnkRep} (without ensemble)}&32.7\\
\multicolumn{2}{c}{\newcite{luong2015addressing}, {\it UnkRep} (with ensemble)}&37.5\\\hline\hline
 \end{tabular}
\caption{BLEU scores from different models for on WMT'14 English-French results. Google
is the LSTM-based model proposed in Sutskever {\em et al.} (2014). Luong {\em et al.} (2015) is the extension of Google models with unknown token replacements. }
\label{English-to-French}
\end{table*}

\subsection{Training p(y) from Monolingual Data}
We respectively  train single-layer LSTM recurrent models with 500 units 
 for German and French
 using monolingual data. 
 We  News Crawl corpora from WMT13\footnote{\url{http://www.statmt.org/wmt13/translation-task.html}} as additional training
data to train monolingual language models. 
We used a subset of the original dataset which 
 roughly contains 50-60 millions sentences. 
  Following  \cite{gulcehre2015using,sennrich2015improving}, we remove sentences with more than $10\%$ Unknown words based on the vocabulary constructed using parallel datasets. 
  We adopted similar protocols as we trained \sts models, such as gradient clipping and mini batch. 
\subsection{English-German Results}
We reported progressive performances as we add in more features for reranking. 
Results for different models on WMT2014 English-German translation task are shown in Figure \ref{tab1}.
Among all the features, reverse probability from mutual information (i.e., p(x$|$y)) yields the most significant performance boost, +1.4 and +1.1 
for standard \sts models without and with unknown word replacement, +0.9 for attention models\footnote{Target length has long proved to be one of the most important features in phrase based MT due to the BLEU score's significant sensitiveness to target lengths. However, here we do not observe as large performance boost here as in phrase based MT. This is due to the fact that during decoding, target length has already been strictly constrained. As described in 4.1, we only consider candidates of lengths between 0.75 and 1.5 times that of the source.}. 
In line with \cite{gulcehre2015using,sennrich2015improving}, we observe consistent performance boost introduced by language model. 

We see
the benefit from our diverse N-best list by comparing  {\it mutual+diversity} models with {\it diversity} models.
On top of the improvements from standard beam search due to reranking, 
the {\it diversity} models introduce additional gains of +0.4, +0.3 and +0.3, 
leading the total gains roughly up to +2.6, +2.6, +2.1 for different models.  
The unknown token replacement technique yields significant gains,
in line with observations from \newcite{jean2014using,luong2015effective}. 

We compare our English-German system with various others:
(1) The end-to-end 
neural
MT system from \newcite{jean2014using} using a large vocabulary size.
(2) Models from \newcite{luong2015effective} that combines different attention models. 
For the models described in \cite{jean2014using} and \cite{luong2015effective},
we reprint their results from both the single model setting
and the {\it ensemble} setting, which a set of 
(usually 8) neural models that differ in  random initializations and the order of minibatches are trained, the combination of which jointly contributes in the decoding process.
The {\it ensemble}  procedure is known to  result in improved performance
\cite{luong2015effective,jean2014using,sutskever2014sequence}.

Note that the reported results from  the
standard \sts models and attention models in Table 1
(those without considering mutual information) are 
from models  identical in structure to the corresponding
models described in \cite{luong2015effective}, and
achieve similar performances (13.2 vs 14.0 for standard \sts models and 20.5 vs 20.7 for attention models). 
Due to time and computational constraints, we did not implement an ensemble mechanism,
making our results incomparable to the ensemble mechanisms in these papers.

\subsection{French-English Results}

Results from the WMT'14 French-English
datasets  are shown in Table~\ref{English-to-French}, along with results
reprinted from \newcite{sutskever2014sequence,luong2015addressing}.
We again observe that applying mutual information
yields better performance than the corresponding standard neural MT models.

Relative to the English-German dataset, the English-French translation task shows
a larger gap between our new
model and vanilla models where reranking information  is not considered;
our models respectively yield up to +3.2, +2.6, +2.7 boost in BLEU compared to
standard neural models without and with unknown word replacement,
and Attention models.

\section{Discussion}

In this paper, we introduce a new objective for neural MT
based on the mutual dependency between the source and target sentences,
inspired by recent work in neural conversation generation \cite{li2015diversity}.
We build an approximate implementation of our model using reranking, and then
to make reranking more powerful we introduce a new decoding
method that promotes diversity in the first-pass N-best list. 
On English$\rightarrow$French and English$\rightarrow$German
translation tasks, we show that the neural machine
translation models trained using the proposed
method perform better than corresponding standard models,
and that both the mutual information objective and the
diversity-increasing decoding methods contribute to the performance boost..

The new models come with the advantages of easy implementation with sources and targets interchanged,
and of offering a general solution that can be integrated into any
neural generation models with minor adjustments. 
Indeed, our diversity-enhancing decoder can be applied to 
generate more diverse N-best lists for any NLP reranking task.
Finding a way to introduce mutual information based decoding directly into a first-pass
decoder without reranking naturally constitutes our future work. 

\bibliographystyle{acl2012}
\bibliography{MMI_MT}

\end{document}